\documentclass[lettersize,journal]{IEEEtran}
\usepackage{amsmath,amsfonts}
\usepackage{algorithmic}
\usepackage{algorithm}
\usepackage{array}
\usepackage[caption=false,font=normalsize,labelfont=sf,textfont=sf]{subfig}
\usepackage{textcomp}
\usepackage{stfloats}
\usepackage{url}
\usepackage[dvipsnames]{xcolor}
\usepackage{hyperref}
\usepackage{verbatim}
\usepackage{graphicx}
\usepackage{cite}
\usepackage{multirow}
\usepackage[flushleft]{threeparttable}
\begin{document}

\title{Fusion-based Few-Shot Morphing Attack Detection and Fingerprinting}

\author{Na Zhang, Shan Jia, Siwei~Lyu,~\IEEEmembership{Fellow,~IEEE}
        and Xin~Li,~\IEEEmembership{Fellow,~IEEE}
\thanks{Na Zhang and Xin Li are with Lane Department of Computer Science and Electrical Engineering, Morgantown, WV 26506-6109. Shan Jia and Siwei Lyu are in the Department of Computer Science and Engineering at the State University of New York at Buffalo, Buffalo, NY14260.}
}

\markboth{Journal of \LaTeX\ Class Files,~Vol.~14, No.~8, Oct~2022}%
{Shell \MakeLowercase{\textit{et al.}}: A Sample Article Using IEEEtran.cls for IEEE Journals}


\maketitle

\begin{abstract}
The vulnerability of face recognition systems to morphing attacks has posed a serious security threat due to the wide adoption of face biometrics in the real world. Most existing morphing attack detection (MAD) methods require a large amount of training data and have only been tested on a few predefined attack models. The lack of good generalization properties, especially in view of the growing interest in developing novel morphing attacks, is a critical limitation with existing MAD research. To address this issue, we propose to extend MAD from supervised learning to few-shot learning and from binary detection to multiclass fingerprinting in this paper. Our technical contributions include: 1) We propose a fusion-based few-shot learning (FSL) method to learn discriminative features that can generalize to unseen morphing attack types from predefined presentation attacks; 2) The proposed FSL based on the fusion of the PRNU model and Noiseprint network is extended from binary MAD to multiclass morphing attack fingerprinting (MAF). 3) We have collected a large-scale database, which contains five face datasets and eight different morphing algorithms, to benchmark the proposed few-shot MAF (FS-MAF) method. Extensive experimental results show the outstanding performance of our fusion-based FS-MAF. The code and data will be publicly available at \textcolor{red}{\url{https://github.com/nz0001na/mad_maf}}.
\end{abstract}


\section{Introduction}
\label{sec:intro}
\par With the fast development of deep learning techniques, face recognition systems (FRSs) have become a popular technique for identifying and verifying people due to the ease of capturing biometrics from the face. In our daily lives, one of the most relevant applications of FRS is the Automatic Border Control system, which can quickly verify the identity of a person with his electronic machine-readable travel document (eMRTD) \cite{icao20159303} by comparing the face image of the traveler with a reference in the database. Although high-accuracy FRS can effectively distinguish an individual from others, it is vulnerable to adversarial attacks that conceal the real identity. Recent research found that attacks based on morphed faces \cite{ferrara2014magic,scherhag2017vulnerability} pose a serious security risk in various applications. 

\begin{figure*}[t]
\centering
\includegraphics[width=.75\linewidth]{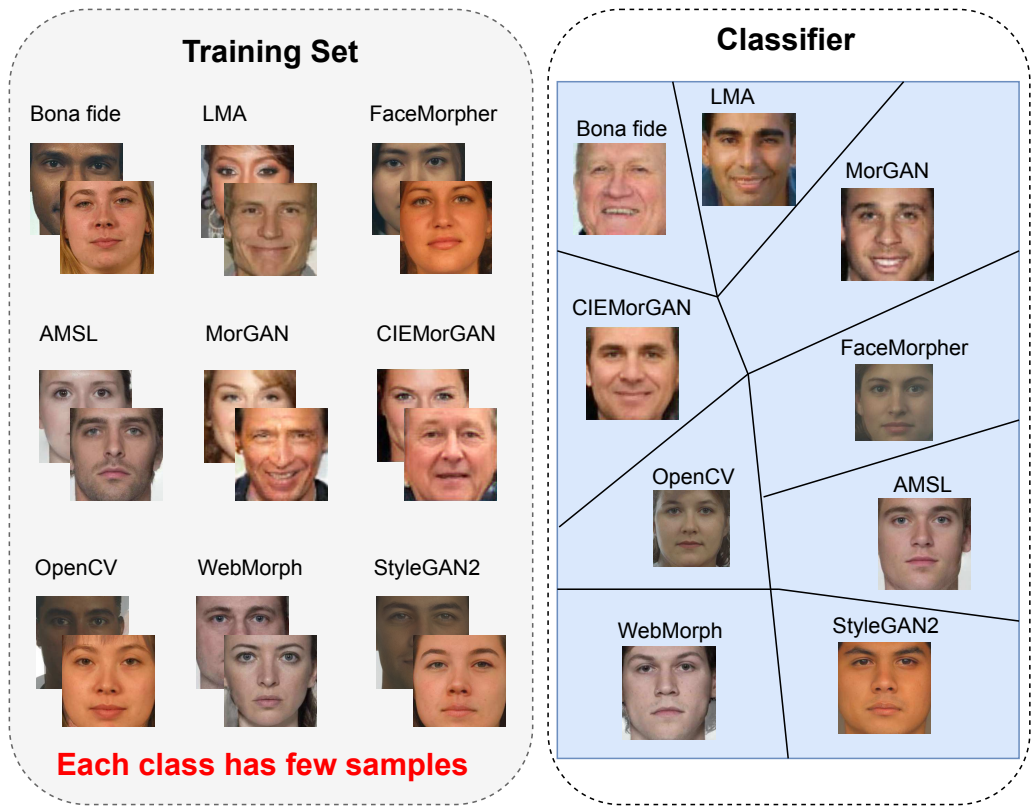}
\caption{Few-shot learning for morphing attack fingerprinting (MAF), a multiclass extension of MAD. Each class (morphing attack model) of the training set contains a few examples. After training, the model can classify unseen test samples for each class.}
\label{fig:MAF}
\end{figure*}

\par Morphing attacks were first introduced in 2014 \cite{ferrara2014magic}. The morphed face is combined by two or more bona fide faces, and it was shown that commercial face recognition software tools are highly vulnerable to such attacks. In a further study \cite{ferrara2016effects}, the authors showed that the images of morphed faces are realistic enough to fool human examiners. With the emergence of face morphing generation techniques \cite{gimp,damer2018morgan,zhang2020mipgan,karras2019style} and numerous easy-to-use face morphing softwares (e.g., MorphThing \cite{morphthing}, 3Dthis Face Morph \cite{3dmorph}, Face Swap Online \cite{faceswap}, Abrosoft FantaMorph \cite{fanta}, FaceMorpher \cite{morpher}), there is an imminent need to protect FRS security by detecting morphing attacks \cite{scherhag2019face}. 

Some morphing attack detection (MAD) approaches have been developed since 2018 (for a recent review, see \cite{venkatesh2021face}). They can be categorized into two types: single image-based (S-MAD) and differential image-based (D-MAD) \cite{raja2020morphing}. The deep face representation for D-MAD has been studied in \cite{scherhag2020deep}; existing S-MAD methods can be further classified into two subtypes \cite{venkatesh2020face}: model-based (using handcraft characteristics) and deep learning-based. Noise-based photo-response non-uniformity (PRNU) methods \cite{debiasi2018prnu,debiasi2018prnu2,scherhag2019detection,zhang2018face} represent the former subtype due to its popularity and outstanding performance. Originally proposed for camera identification, PRNU turns out to be useful for detecting the liveness of face photos. For the latter subtype, Noiseprint \cite{cozzolino2018noiseprint} used a CNN to learn the important features, with the objective of improving detection performance and supporting fingerprinting applications. 

\par Despite rapid progress, existing MAD methods are often constructed on a small training dataset and a single modality, which makes them lacking good generalization properties \cite{raja2020morphing,venkatesh2020face}. The performance of existing MAD methods might be satisfactory for predefined morphing attack models, but degrades rapidly when deployed in the real world facing newly evolved attacks. Although it is possible to alleviate this problem by fine-tuning the existing MAD model, the cost of collecting labeled data for every new morphing attack is often formidable. Furthermore, we argue that MAD alone is not sufficient to meet the demand for increased security risk facing FRS. A more aggressive countermeasure than MAD to formulate the problem of morphing attack fingerprinting (MAF), that is, we aim at a multiclass classification of morphing attack models, as shown in Fig. \ref{fig:MAF}.

\par Based on the above observations, we propose to formulate MAF as a few-shot learning problem in this paper. Conventional few-shot learning (FSL) \cite{snell2017prototypical} learns the knowledge from a few examples of each class and predicts the class label of the new test samples. Similarly, we train the detector using data from both predefined models and new attack models (only a few samples are required) to predict unknown new test samples. This task is named the few-shot MAD (FS-MAD) problem. Unlike existing MAD research, few-shot MAF (FS-MAF) aims at learning general discriminative features, which can be generalized from predefined to new attack models. The problem of few-shot MAF is closely related to camera identification (ID) \cite{lukas2006digital}, camera model fingerprinting \cite{cozzolino2018noiseprint}, and GAN fingerprinting (a.k.a. model attribution \cite{yu2019attributing}) in the literature. The main contributions of this paper are summarized below.

$\bullet$ Problem formulation of few-shot learning for MAD/MAF. We challenge the widely accepted assumptions of the MAD community, including the NIST's FRVT MORPH competition. The generalization property of MAD/MAF methods will be as important as the optimization of recognition accuracy. 

$\bullet$ Feature-level fusion for MAD applications. Although both PRNU and Noiseprint have shown promising performance in camera identification applications, no one has demonstrated their complementary nature in the open literature. We believe that this work is the first to combine them through feature-level fusion and to study the optimal fusion strategy.

$\bullet$ Design a fusion-based FSL method with adaptive posterior learning (APL) for MAD/MAF. By adaptively combining the most surprising observations encountered by PRNU and Noiseprint, we can achieve a good generalization property by optimizing the performance of FS-MAD/FS-MAF at the system level. 

$\bullet$ Construction of a large-scale benchmark dataset to support MAD/MAF research. More than 20,000 images with varying spatial resolution have been collected from various sources. Extensive experimental results have justified the superior generalization performance of FS-MAD and FS-MAF over all other competing methods.

\section{Related Work}
\label{related}

\subsection{Morphing Attack Detection (MAD)}
\noindent \textbf{Model-based S-MAD}. Residual noise feature-based methods are designed to analyze pixel discontinuity, which may be greatly affected by the morphing process. Generally, noise patterns are extracted by subtracting the given image from a denoised version of the same image using different models, such as the deep multiscale context aggregate network (MS-CAN) \cite{venkatesh2020detecting}. The most popular should be sensor noise patterns, such as PRNU. Recently, both PRNU-based \cite{zhang2018face,debiasi2018prnu,debiasi2018prnu2,scherhag2019detection} and scale-space ensemble approaches \cite{raja2020morphing,raja2017transferable} have been studied.

\noindent \textbf{Learning-based S-MAD}. Along with rapid advances in deep learning, many methods have considered the extraction of deep learning features for detection. The use of a convolutional neural network (CNN) has reported promising results \cite{8897214}. Most works are based on pre-trained networks and transfer learning. Commonly adopted deep models contain AlexNet \cite{krizhevsky2012imagenet}, VGG16 \cite{simonyan2014very}, VGG19 \cite{simonyan2014very,raja2017transferable}, GoogleNet \cite{szegedy2015going}, ResNet \cite{he2016deep}, etc. In addition, several self-design models were also proposed. More recently, a deep residual color noise pattern was proposed for MAD in \cite{venkatesh2019morphed}; and an attention-based deep neural network (DNN) \cite{aghdaie2021attention} was studied, focusing on the salient regions of interest (ROI) that have the most spatial support for the morph detector decision function.

\noindent\textbf{Learning-based D-MAD}. The presented D-MAD methods mainly focus on feature differences and demorphing. For feature difference-based methods, features of the suspected image and the live image are subtracted and further classified. Texture information, 3D information, gradient information, landmark points, and deep feature information (ArcFace \cite{scherhag2020deep}, VGG19 \cite{seibold2020accurate}) are the most popular features used. The authors in \cite{scherhag2018detecting} computed distance-based and angle-based features of landmark points for analysis. In \cite{singh2019robust}, a robust method using diffuse reflectance in a deep decomposed 3D shape was proposed. Fusion methods were commonly adopted by concatenating hand-crafted Local Binary Pattern Histogram (LBPH) and transferable deep CNN features \cite{damer2019multi}, or concatenating feature vectors extracted from texture descriptors, keypoint extractors, gradient estimators and deep neural networks \cite{scherhag2018towards}. More recently, a discriminative DMAD method in the wavelet subband domain was developed to discern the disparity between a real and a morphed image.

\subsection{Few-Shot Learning (FSL)}
Few-shot learning addresses the challenge with the generalization property of deep neural networks, i.e., how can a model quickly generalize after only seeing a few examples from each class? Early approaches include meta-learning models \cite{ravi2016optimization} and deep metric learning techniques \cite{snell2017prototypical}. More recent advances have explored new directions such as the relation network \cite{sung2018learning}, meta-transfer learning \cite{sun2019meta}, adaptive posterior learning (APL) \cite{ramalho2019adaptive}, and cluster-based object seeker with shared object concentrator (COSOC) \cite{luo2021rectifying}.

\subsection{Camera and Deepfake Fingerprinting}
\par PRNU, as a model-based device fingerprint, has been used to perform multiple digital forensic tasks, such as device identification \cite{cozzolino2020combining}, device linking \cite{salazar2021evaluation}, forgery localization \cite{lin2020prnu}, detection of digital forgeries \cite{lugstein2021prnu}. It can find any type of forgery, irrespective of its nature, since the lack of PRNU is seen as a possible clue of manipulation. Furthermore, PRNU-based MAD methods \cite{debiasi2018prnu,debiasi2018prnu2,scherhag2019detection,zhang2018face} also confirm the usefulness of the sensor fingerprint in MAD.
In recent years, PRNU has been applied successfully in MAD \cite{debiasi2018prnu2,debiasi2018prnu,scherhag2019detection}. The method in \cite{debiasi2018prnu2} shows that region-based PRNU spectral analysis reliably detects morphed face images, while it fails if image post-processing is applied to generated morphs. Based on previous work, a PRNU variance analysis was performed in \cite{debiasi2018prnu}. It focused on local variations of face images, which can be useful as a reliable indicator for image morphing. The work in \cite{scherhag2019detection} proposed an improved version of the scheme based on the previous PRNU variance analysis in image blocks. Another work \cite{marra2019gans} showed that each GAN model leaves a specific fingerprint in the generated images, just as the PRNU traces left by different cameras in real-world photos.

\begin{figure*}[!t]
\centering
\includegraphics[width=1.0\linewidth]{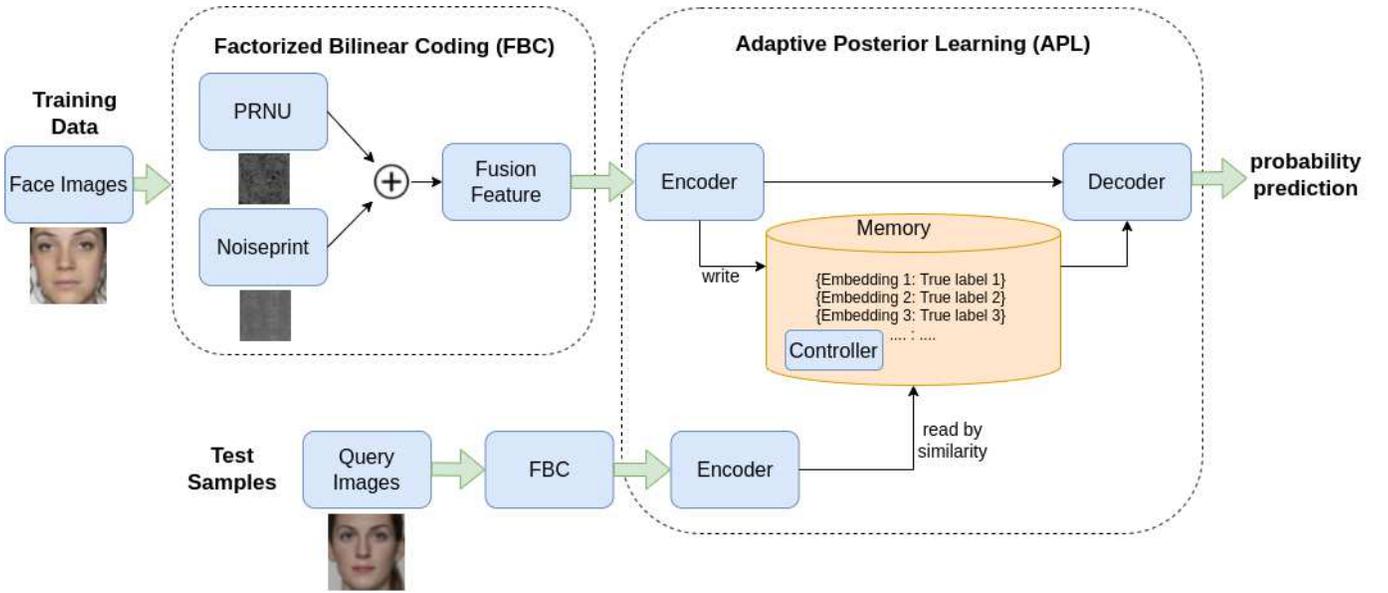}
\vspace{-0.75cm}
\caption{An overview of the proposed system (FBC-APL) for few-shot MAF (FS-MAF). It consists of factorized bilinear coding (FBC) and adaptive posterior learning (APL) modules. The output contains the probability that the input image will be classified into one of the known morphing models.}
\label{fig:pipeline}
\end{figure*}

\section{Methodology}
\label{funda}

Morphing attack fingerprinting (MAF) refers to the multiclass generalization of the existing binary MAD problem. In addition to detecting the presence of morphing attacks, we aim at finer-granularity classification about the specific model generating the face morph. It is hypothesized that different attack models inevitably leave fingerprints in morphed images (conceptually similar to the sensor noise fingerprint left by different camera models \cite{lukas2006digital}).
Fig. \ref{fig:pipeline} shows the overall system consisting of two stages: feature fusion through factorized bilinear coding (FBC) and few-shot learning (FSL) for MAF. We will first elaborate on fusion-based MAD in detail and then discuss the extension to few-shot MAF.

\subsection{Fusion-based Single-Image MAD}
\par Noise is often embedded in the image data during acquisition or manipulation. The uniqueness of the noise pattern is determined by the physical source or an artificial algorithm, which can be characterized as a statistical property to reveal the source of the noise \cite{popescu2004statistical}. The noise of the sensor pattern was first used for the MAD task by performing a facial quantification statistics analysis, which confirmed its effectiveness \cite{zhang2018face}. Here, we consider two types of sensor noise patterns: Photo Response Non-Uniformity (PRNU) \cite{fridrich2009digital} and Noiseprint \cite{cozzolino2018noiseprint}.

\noindent \textbf{Photo Response Non-Uniformity (PRNU)}. PRNU originates from slight variations between individual pixels during photoelectric conversion in digital image sensors \cite{lukas2006digital}. Different image sensors embed this weak signal into acquired images as a unique signature. Although the weak signal itself is mostly imperceptible to the human eye, its uniqueness can be characterized by statistical techniques and exploited by sophisticated fingerprinting methods such as PRNU \cite{fridrich2009digital}. 
This systemic and individual pattern, which plays the role of a sensor fingerprint, has proven robust to various innocent image processing operations such as JPEG compression. Although PRNU is stochastic in nature, it is a relatively stable component of the sensor over its lifetime. 

PRNU has been widely studied in camera identification because it is not related to image content and is present in every image acquired by the same camera. Most recently, PRNU has been proposed as a promising tool for detecting morphed face images \cite{debiasi2018prnu,debiasi2018prnu2}.
The spatial feature of PRNU can be extracted using the approach presented by Fridrich \cite{fridrich2009digital}. For each image $I$, the residual noise ${W}_{I}$ is estimated as described in Equation \eqref{eq1}:
\begin{equation}
\label{eq1}
\vspace{-0.1in}
{W}_{I} = I - F(I) 
\end{equation}
where $F$ is a denoising function that filters the noise from the sensor pattern. The clever design of the mapping function $F$ (e.g., wavelet-based filter \cite{lukas2006digital}) makes PRNU an effective tool for various forensic applications.

\noindent \textbf{Noiseprint}. Unlike model-based PRNU, data-driven or learning-based methods tackle the problem of camera identification by assuming the availability of training data. Instead of mathematically constructing unique signatures, Noiseprint \cite{cozzolino2018noiseprint} attempts to learn the embedded noise pattern from the training data. A popular learning methodology adopted by Noiseprint is to construct a Siamese network \cite{bertinetto2016fully}. The Siamese network is trained with pairs of image patches that come from the same or different cameras in an unsupervised manner. Similarly to PRNU, Noiseprint has shown clear traces of camera fingerprints. It should be noted that Noiseprint has performed better than PRNU when cropped image patches become smaller, implying the benefit of exploiting spatial diversity \cite{cozzolino2018noiseprint}.

\par To the best of our knowledge, Noiseprint has not been proposed for MAD in the open literature. Existing deep learning-based S-MADs often use pre-trained networks such as VGG-face \cite{raja2020morphing}. Our empirical study shows that morphing-related image manipulation leaves evident traces in Noiseprint, suggesting the feasibility of Noiseprint-based MAD. Moreover, morphed faces are often manipulated across the face, whose spatial diversity can be exploited by cropping image patches using Noiseprint. To justify this claim, Fig. \ref{fig:featurefig} (d) presents the Noiseprint comparison between bona fide and morphed faces averaged over 1,000 examples. Visual inspection clearly shows that the areas around the eyes and nose have more significant (bright) traces than the bona fide faces. In contrast, Fig. \ref{fig:featurefig} (c) shows the comparison of the extracted PRNU patterns with the same experimental setting. Similar visual differences between bona fide and morphed faces can be observed; more importantly, PRNU and Noiseprint demonstrate complementary patterns (low vs. high frequency) begging for fusion.

\noindent \textbf{Feature Fusion Strategy}. Fusion methods are usually based on multiple feature representations or classification models. Taking advantage of diversity, the strategy of combining classifiers \cite{kittler1998combining} has shown improved recognition performance compared to single-mode approaches. Recent work has shown that fusion methods based on Dempster-Shafer theory can improve the performance of face morphing detectors \cite{makrushin2019dempster}. However, previous work \cite{makrushin2019dempster} only considered ensemble models of the scale space and pre-trained CNN models. For the first time, we propose to combine PRNU and Noiseprint using a recently developed similarity-based fusion method, called factorized bilinear coding (FBC) \cite{gao2020revisiting}.

FBC is a sparse coding formulation that generates a compact and discriminative representation with substantially fewer parameters by learning a dictionary $\boldsymbol{B}$ to capture the structure of the entire data space. It can preserve as much information as possible and activate as few dictionary atoms as possible. Let $\boldsymbol{x}_i$, $\boldsymbol{y}_j$ be the two features extracted from PRNU and Noiseprint, respectively. The key idea behind FBC is to encode the extracted features based on sparse coding and to learn a dictionary $\boldsymbol{B}$ with $k$ atoms by matrix factorization. Specifically, the sparsity FBC opts to encode the two input features $(\boldsymbol{x}_i, \boldsymbol{y}_j)$ in the FBC code $\boldsymbol{c}_v$ by solving the following optimization problem:

\begin{equation}
\underset{{{\boldsymbol{c}}_{v}}}{\mathop{\min }}\,\bigg|\bigg|{{\boldsymbol{x}}_{i}}\boldsymbol{y}_{j}^{\top}-\sum\limits_{l=1}^{k}{c_{v}^{l}}{{\boldsymbol{U}}_{l}}\boldsymbol{V}_{l}^{\top}\bigg|{{\bigg|}^{2}}+\lambda||{{\boldsymbol{c}}_{v}}|{{|}_{1}}
\end{equation}
where $\lambda$ is a trade-off parameter between the reconstruction error and the sparsity. The dictionary atom $b_l$ of $\boldsymbol{B}$ is factorized into $\boldsymbol{U}_{l}\boldsymbol{V}_{l}^{\top}$ where 
    $\boldsymbol{U}_{l}$ and $\boldsymbol{V}_{l}^{\top}$ are low-rank matrices. The $l_1$ norm $|| \cdot ||_1$ is used to impose the sparsity constraint on $\boldsymbol{c}_{v}$. In essence, the bilinear feature $\boldsymbol{x}_{i}\boldsymbol{y}_{j}^{\top}$ is reconstructed by $\sum\limits_{l=1}^{k}{c_{v}^{l}} \boldsymbol{U}_{l}\boldsymbol{V}_{l}^{\top}$
with $\boldsymbol{c}_{v}$ being the FBC code and $c_v^l$ representing the $l$-th element of $\boldsymbol{c}_{v}$.

\begin{figure}[t]
\centering
\includegraphics[width=1.0\linewidth]{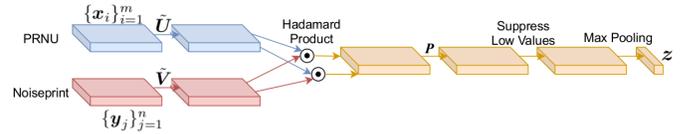}
\vspace{-0.65cm}
\caption{The architecture of the FBC module to combine PRNU and Noiseprint. $\tilde{\boldsymbol{U}}$ and $\tilde{\boldsymbol{V}}$ replace $\boldsymbol{U}$ and $\boldsymbol{V}$ to avoid numerically unstable matrix inversion operations; $\boldsymbol{P}$ is a fixed binary matrix.}
\label{fig:fbc}
\end{figure}

This optimization can be solved using well-studied methods such as LASSO \cite{tibshirani1996regression}. With two groups of features $\{\boldsymbol{x}_i\}_{i=1}^m$ and $\{\boldsymbol{y}_j\}_{j=1}^n$ at our disposal, we first calculate all FBC codes $\{\boldsymbol{c}_v\}_{v=1}^N$ and then fuse them by the operation $max$ to achieve global representation $\boldsymbol{z}$: 
\begin{equation}
    \boldsymbol{z}=max\left\{\boldsymbol{c}_{v}\right\}_{i=1}^{N}.
    \label{eq:3}
\end{equation}
The entire FBC module is shown in Fig. \ref{fig:fbc}.

\subsection{Few-shot learning for Morphing Attack Fingerprinting}
\par Based on the FBC-fused feature $\boldsymbol{z}$, we construct a few-shot learning module as follows. Inspired by recent work on adaptive posterior learning (APL) \cite{ramalho2019adaptive}, we have redesigned the FSL module to adaptively select feature vectors of any size (e.g., FBC-fused feature) as input. This newly designed module consists of three parts: an encoder, a decoder, and an external memory store. The encoder is used to generate a compact representation for the incoming query data; the memory saves the previously seen representation by the encoder; the decoder aims at generating a probability distribution over targets by analyzing the query representation and pairwise data returned from the memory block. Next, we will elaborate on the design of these three components.

\noindent\textbf{Encoder}. The encoder can convert input data of any size to a compact embedding with low dimensionality. It is implemented by a convolutional network, which is composed of a single first convolution to map the input to 64 feature channels, followed by 15 convolutional blocks. Each block is made up of a batch normalization step, followed by a ReLU activation and a convolutional layer with kernel size 3. For every three blocks (one combo), the convolution contains a stride 2 to down-sample the image. All layers have 64 features. Finally, the feature is flattened to a 1D vector and passed through Layer Normalization, generating a 64-dimensional embedding as an encoded representation.

\begin{figure*}[t]
\centering
\includegraphics[width=1.0\linewidth]{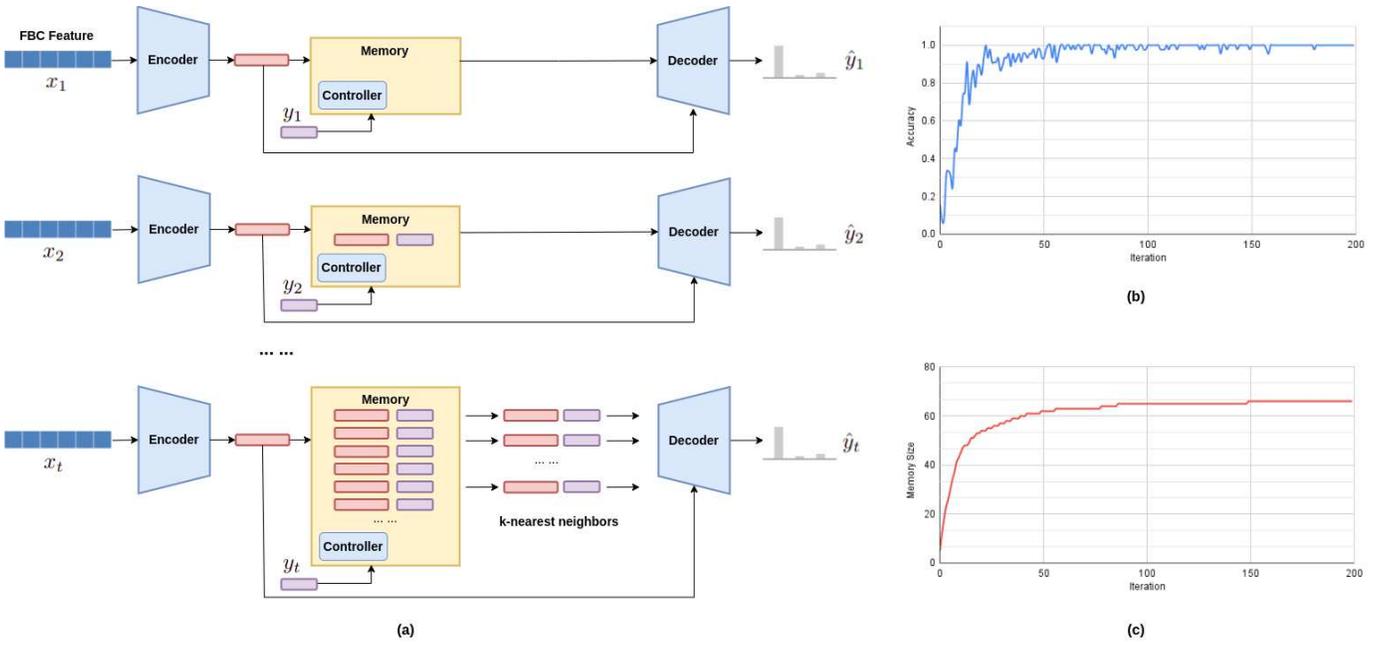}
\vspace{-0.4cm}
\caption{(a) APL training procedure for iterations. We train the APL module on a sequence of episodes ($x_t$, $y_t$), where $x_t$ is the FBC feature and $y_t$ is the true label. At first, the memory is empty; at each iteration, a batch of samples is fed to the module, and a prediction is made. Cross-entropy loss L($\hat{y}_t$, $y_t$) is calculated and a gradient update step is performed to minimize the loss in that batch alone. The loss is also fed to the memory controller so that the network can decide whether to write to memory. (b) and (c) show the behavior of the accuracy and memory size in a 9-class training scenario. APL stops writing to memory after having about 7 examples per class for classification.}
\label{fig:fscnn}
\end{figure*}

\noindent\textbf{Memory}. The external memory store is a database to store experiences. It is key-value data. Each row represents the information for one data point. Each column is decomposed into an embedding (encoded representation) and a true label. The memory store is managed by a controller that decides which embeddings can be written into the memory while at the same time tries to minimize the amount of written embeddings. During the writing process, a quantity metric surprise is defined. The higher the probability that the model assigns to the true class correctly, the less surprised it will be. If the confidence in the prediction in the correct class is smaller than the probability assigned by a uniform prediction, the embedding should be written into memory. During the querying process, the memory is queried for the k-nearest-neighbors of the embeddings of queries from the encoder. The distance metric used to calculate the proximity between points is an open choice, and here we use two types (euclidean distance and cosine distance). Both the full-row data for each of the neighbors and query embeddings are concatenated and fed to the decoder. 

\begin{figure}
\centering
\includegraphics[width=0.8\linewidth]{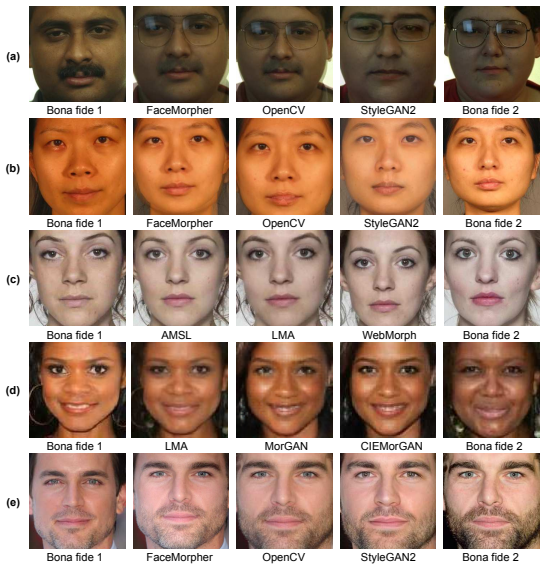}
\caption{Face samples in five merged datasets. (a) FERET-Morphs (bona fide faces come from FERET \cite{feret}), (b) FRGC-Morphs (bona fide faces come from FRGC V2.0 \cite{frgc}), (c) FRLL-Morphs (bona fide faces come from Face Research Lab London Set (FRLL) \cite{amslraw}), (d) CelebA-Morphs (bona fide faces come from CelebA \cite{liu2015deep}), and (e) Doppelgänger Morphs (bona fide faces come from the Web collection).}
\label{fig:sample}
\end{figure}

\begin{figure*} 
\centering
\includegraphics[width=0.95\linewidth]{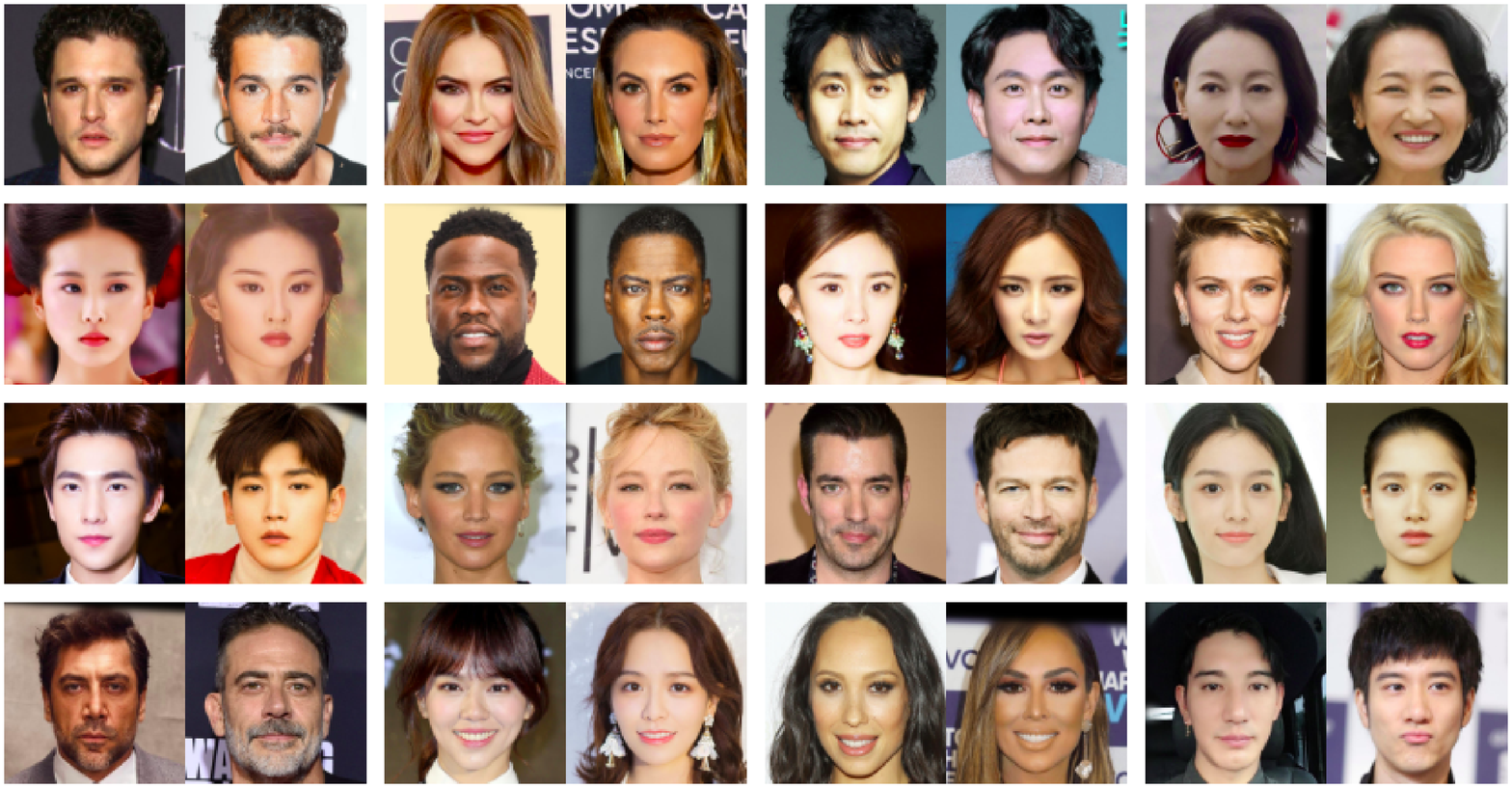}
\caption{Some sample pairs of bona-fide face images from the Doppelgänger dataset (note that these look-alike pairs do not have biological connections).}.
\label{fig:self}
\end{figure*}

\noindent\textbf{Decoder}. The decoder takes the concatenation of query embedding, recalled neighbor embeddings from memory, labels, and distances as input. The architecture is a self-attention-based relational feedforward module. It processes each of the neighbors individually by comparing them with the query and then does a cross-element comparison with a self-attention module before reducing the activations with an attention vector calculated from neighbor distances. The self-attention blocks are repeated five times in a residual manner. The resulting tensors are called activation tensors. In addition, the distances between neighbors and the query are passed through a softmax layer to generate an attention vector, which is summed with the activation tensor on the first axis to obtain the final logit result for classification. The self-attention block comprises a multihead attention layer, a multihead dot product attention (MHDPA) layer \cite{santoro2018relational} for cross-element comparison, and a nonlinear multilayer perceptron (MLP) layer to process each element individually.

\noindent\textbf{Training}. During APL training, as shown in Fig. \ref{fig:fscnn} (a), the query data (that is, the FBC-fused feature vector $\boldsymbol{z}$) are passed through the encoder to generate an embedding, and this representation is used to query an external memory store. At first, the memory is empty; at each training episode, a batch of examples is fed to the model, and a prediction is made. Cross-entropy loss is used to be fed to the memory controller to decide whether to write to memory. After the query is searched in memory, the returned memory contents, as well as the query, are fed to the decoder for classification. Figs. \ref{fig:fscnn} (b) and (c) show the behavior (accuracy and memory size) of APL during a single episode. The accuracy of APL increases as it sees more samples and saturates at some point, indicating that the additional inputs do not surprise the module anymore. In the case of the 9-class classification scenario, we have observed that about 7 examples per class are sufficient to reach performance saturation.

\noindent\textbf{Morphing Attack Fingerprinting}. Both PRNU \cite{lukas2006digital} and Noiseprint \cite{cozzolino2018noiseprint} were originally proposed for the identification of camera models, which is known to be a fingerprint in image forensics. The duality between image generation in the cyber and physical worlds inspires us to extend the existing problem formulation of binary MAD \cite{debiasi2018prnu,debiasi2018prnu2,scherhag2019detection,zhang2018face} into multiclass fingerprinting. Different camera models (e.g., Sony vs. Nikon) are analogous to varying face morphing methods (e.g., LMA \cite{damer2018morgan} vs. StyleGAN2 \cite{karras2020analyzing}); therefore, it is desirable to go beyond MAD by exploring the feasibility of distinguishing one morphing attack from another. Fortunately, the system shown in Fig. \ref{fig:pipeline} easily lends itself to generalization from binary to multiclass classification by resetting the hyperparameters, like the number of classes, the data path for each class, etc. To learn a discriminative FBC feature for fingerprinting, multiclass labeled data for training and testing should be prepared to be fed to the FBC module for retraining. When the FBC feature is available, it will be fed to the APL module for multiclass classification.

\section{Experiments}
\label{result}

\begin{table}[t]
\begin{center}
\small
\caption{The hybrid face morphing benchmark database consists of five image sources and 3-6 different morphing methods for each.}
\label{tab:dbinfo}
\vspace{-0.3cm}
\begin{threeparttable}
\begin{tabular}{ l |l c c}
\hline\noalign{\smallskip}
Database & Subset & \#Images & Resolution \\
\noalign{\smallskip}\hline\noalign{\smallskip}
\multirow{4}{*}{FERET-Morphs} &	bona fide \cite{feret} &	576 & 512x768\\
	& FaceMorpher \cite{sarkar2020vulnerability} &	529 & 512x768\\
	& OpenCV \cite{sarkar2020vulnerability} &529 & 512x768\\
	& StyleGAN2 \cite{sarkar2020vulnerability} & 529 & 1024x1024 \\
\hline
\multirow{4}{*}{FRGC-Morphs} & bona fide \cite{frgc} & 	964 & 1704x2272\\
	& FaceMorpher \cite{sarkar2020vulnerability} &	964 & 512x768\\
	& OpenCV \cite{sarkar2020vulnerability} & 964 & 512x768\\
	& StyleGAN2 \cite{sarkar2020vulnerability} & 964 & 1024x1024\\
\hline
\multirow{7}{*}{FRLL-Morphs} & bona fide \cite{amslraw} & 102+1932 & 413x531\\
	& AMSL \cite{neubert2018extended} & 2175 & 413x531 \\
	& FaceMorpher \cite{sarkar2020vulnerability} &	1222 & 431x513 \\
	& OpenCV \cite{sarkar2020vulnerability}& 1221 & 431x513 \\
	& LMA  &768 & 413x531\\
	& WebMorph \cite{sarkar2020vulnerability} & 1221 & 413x531\\
	& StyleGAN2 \cite{sarkar2020vulnerability} & 1222 & 1024x1024 \\
\hline
\multirow{4}{*}{CelebA-Morphs*} & bona fide \cite{liu2015deep} & 2989 & 128x128 \\
	& MorGAN \cite{damer2018morgan}& 1000 & 64x64\\
	& CIEMorGAN \cite{damer2019realistic} & 1000 & 128x128 \\
	& LMA \cite{damer2018morgan} & 1000 & 128x128 \\
\hline
\multirow{4}{*}{Doppelgänger} & bona fide & 306 & 1024x1024 \\
	& FaceMorpher &	150 & 1024x1024 \\
	& OpenCV &	153 & 1024x1024\\
	& StyleGAN2	& 153 & 1024x1024 \\
\noalign{\smallskip}\hline
\end{tabular}
\begin{tablenotes}
\small
\item * means only the cropped faces from raw images are used; no facial cropping is used for other datasets. The raw number of bona fide images in FRLL-Morphs is 102. Based on the raw faces, data augmentation is implemented to obtain extra 1932 images. 
\end{tablenotes}
\end{threeparttable}
\vspace{-0.2in}
\end{center}
\end{table}

\subsection{Large-scale Morphing Benchmark Dataset}
\noindent \textbf{Benchmark Dataset Description.} To simulate the amount and distribution of data in real-world applications, we have combined five datasets to build a large-scale evaluation benchmark for detecting and fingerprinting few-shot morphing attacks. It contains four publicly available datasets, namely, FERET-Morphs \cite{feret,sarkar2020vulnerability}, FRGC-Morphs \cite{frgc,sarkar2020vulnerability}, FRLL-Morphs \cite{amslraw,neubert2018extended,sarkar2020vulnerability}, and CelebA-Morphs \cite{liu2015deep,damer2018morgan,damer2019realistic}. We also generated a new dataset with high-resolution faces collected from the Web, named Doppelgänger Morphs, which contains morphing attacks from three algorithms and satisfies the so-called Doppelgänger constraint \cite{rottcher2020finding} (that is, look-alike faces without biological connections, refer to Fig. \ref{fig:self}). A total of more than 20,000 images (6,869 bona fide faces and 15,764 morphed faces) have been collected, as shown in Table \ref{tab:dbinfo}. Eight morphing algorithms are involved, including five landmark-based methods, OpenCV \cite{opencv}, FaceMorpher \cite{facemorpher}, LMA \cite{damer2018morgan}, WebMorph \cite{webmorph}, and AMSL \cite{neubert2018extended}, and three adversarial generative networks based, including MorGAN \cite{damer2018morgan}, CIEMorGAN \cite{damer2019realistic}, and StyleGAN2 \cite{karras2020analyzing}. Fig. \ref{fig:sample} provides some cropped face samples with real faces and morphed faces from different morphing algorithms in these five datasets. To the best of our knowledge, this is one of the largest and most diverse face morphing benchmarks that can be used for MAD and MAF evaluations.

\noindent \textbf{Evaluation Protocols.}
Based on the large-scale dataset collected for few-shot MAD and MAF benchmarks, we have designed the evaluation protocols for each task as follows:

$\bullet$ Protocol FS-MAD (few-shot MAD). This protocol is designed for the few-shot binary classification (bona fide/morphed). Training data comes from predefined types and a few (1 or 5) samples per new type. The test data come from new types. Here, the predefined types in our experiment contain five types of morphing results generated by FaceMorpher \cite{facemorpher}, OpenCV \cite{opencv}, WebMorph \cite{webmorph}, StyleGAN2 \cite{karras2020analyzing}, and AMSL \cite{neubert2018extended}, and their corresponding bona fide faces. Faces of these types are from the FERET-Morphs, FRGC-Morphs, FRLL-Morphs, and Doppelgänger-Morphs datasets. The morphing faces generated by LMA \cite{damer2018morgan}, MorGAN \cite{damer2018morgan}, and CIEMorGAN \cite{damer2019realistic}, and their corresponding bona fide faces, are treated as new types. Faces of these types are from the CelebA-Morphs dataset.

$\bullet$ Protocol FS-MAF (few-shot MAF). This protocol is designed for multiclass fingerprint classification on the hybrid large-scale benchmark and for five separate morph datasets. Each morphing type and bona fide type are treated as different categories, namely FERET-Morphs, FRGC-Morphs, CelebA-Morphs, and Doppelgänger datasets all with 4 classes, FRLL-Morphs with 7 classes, and the hybrid with 9 classes. For each data set, the data are split according to the rule of 8: 2. Training data consist of 1 and 5 images per class for 1 shot and 5-shot learning, respectively. The testing data contains non-overlapping data with the training in each dataset. To reduce the bias of the imbalanced distribution of the data, a similar number of faces is maintained for each class in each test set. 

\begin{table}[!t]
\begin{center}
\caption{Traditional MAD performance (Accuracy-\%) comparison of different feature-level fusion methods. NP - Noiseprint; CN - Concatenation; CC - Convex Compression; $\bot$ - spatial; $\square$ - spectral.}
\vspace{-0.3cm}
\label{tab:toyexp}
\begin{tabular}{ l c c c c c c }
\hline\noalign{\smallskip}
{Feature} & CN & Sum & Max & CC & FBC (ours) \\
\noalign{\smallskip}\hline\noalign{\smallskip}
PRNU $\bot$+PRNU $\square$ & 83.78 & 84.23 & 83.78 & 84.23 & 84.42 \\
NP $\bot$ + NP $\square$ & 89.19 & 89.64 & 89.64 & 89.64 & 96.40\\
PRNU $\bot$ + NP $\square$ & 89.19 & 89.19 & 89.64 & 89.19 & 89.59\\
PRNU $\square$ + NP $\bot$ & 83.78 & 84.23 & 83.78 & 85.59 & 86.04\\
PRNU $\square$. + NP $\square$ & 86.94 & 85.59 & 85.59 & 86.94 & 84.68 \\
PRNU $\bot$ + NP $\bot$ & \textbf{91.44} & \textbf{91.89} & \textbf{91.89} & \textbf{94.59} & \textbf{96.85}\\
\noalign{\smallskip}\hline
\end{tabular}
\end{center}
\end{table}

\begin{table}[!t]
\begin{center}
\caption{Performance (\%) comparison of few-shot MAD. Accu. - Accuracy.}
\vspace{-0.3cm}
\label{tab:madfs}
\resizebox{.95\linewidth}{!}{
\begin{tabular}{ l |c c c |c c c}
\hline\noalign{\smallskip}
 & \multicolumn{3}{c}{1-shot} & \multicolumn{3}{|c}{5-shot} \\
Method & Accu. & D-EER & ACER & Accu. &	D-EER &	ACER \\
\noalign{\smallskip}\hline\noalign{\smallskip}
Xception \cite{chollet2017xception} & 66.50 & 32.50 & 33.50 & 73.25 & 27.00 & 26.75 \\
MobileNetV2 \cite{sandler2018mobilenetv2} & 67.00 & 36.50 & 33.00 & 71.25 & 29.00 & 28.75 \\
NasNetMobile \cite{zoph2018learning} & 59.00 & 40.50 & 41.00 & 66.25 & 35.00 & 33.75 \\
DenseNet121 \cite{huang2017densely} &68.25 & 31.50 & 31.75 & 73.50 & 24.50 & 26.50 \\
ArcFace \cite{deng2019arcface} & 58.00 & 41.00 & 42.00 & 62.25 &	37.50 & 37.75 \\
\hline
Raghavendra. et al. \cite{raghavendra2017face} & 49.25 & 48.00 & 50.75 & 46.75 & 47.50 & 53.25 \\
MB-LBP \cite{scherhag2020face} & 61.00 & 38.50 & 39.00 & 69.25 & 31.00 & 30.75 \\
FS-SPN \cite{zhang2018face} & 51.50 & 45.00 & 48.50 & 58.25 & 43.50 & 41.75 \\	
Pipeline Footprint \cite{neubert2018reducing} & 54.25 & 44.50 &45.75 &	60.25 &	38.50 &	39.75 \\
PRNU Analysis \cite{debiasi2018prnu} & 56.50 & 57.00 & 43.50 & 64.25 & 66.70 & 35.75 \\
Inception-MAD \cite{damer2022privacy} & 62.00 & 34.50 & 38.00 & 67.75 & 32.50 & 32.25 \\
MixFaceNet-MAD \cite{damer2022privacy} & 76.10 & 27.50 & 28.00 & 82.16 & 24.50 & 24.25 \\
Noiseprint-SVM \cite{cozzolino2018noiseprint} & 53.75 & 50.50 & 46.25 & 61.25 & 38.50 & 38.75 \\
\hline
Meta-Baseline \cite{chen2021meta} & 60.45 & - & - & 71.38 & - & -  \\
COSOC \cite{luo2021rectifying} & 66.89 & -&-& 74.54 &-&-  \\
\hline
\textbf{FBC-APL} & \textbf{99.25} & \textbf{1.50} & \textbf{0.75} & \textbf{99.75} & \textbf{0.50} & \textbf{0.25} \\
\noalign{\smallskip}  \hline
\end{tabular}}
\end{center}
\end{table}

\subsection{Experimental Settings}
\noindent \textbf{Data Preprocessing}. Dlib face detector \cite{king2009dlib} is used to detect and crop the face region. The cropped face is normalized according to the coordinates of the eye and resized to a fixed size of $270\times270$ pixels. The feature extraction of PRNU and Noiseprint is performed on the processed faces, respectively. The resulting vector dimension for each type of feature is 72,900 ($270\times270$). 

\noindent \textbf{Performance Metrics}. Following previous MAD studies \cite{raja2020morphing,scherhag2020deep}, we report performance using four metrics, including: (1) Accuracy; (2) D-EER; (3) ACER; (4) Confusion Matrix. Detection Equal-Error-Rate(D-EER) is the error rate for which both BPCER and APCER are identical. Average Classification Error Rate (ACER) is calculated by the mean of the APCER and BPCER values. Attack Presentation Classification Error Rate (APCER) reports the proportion of morph attack samples incorrectly classified as bona fide presentation, and the Bona Fide Presentation Classification Error Rate (BPCER) refers to the proportion of bona fide samples incorrectly classified as morphed samples. Both APCER and BPCER are commonly used in previous studies of MAD \cite{raja2020morphing,scherhag2020deep}.

\subsection{Comparison of Feature Extraction and Fusion Strategies}
First, we show the visual comparison of extracted features by different methods.

\begin{figure}[t]
\centering
\includegraphics[width=0.9\columnwidth]{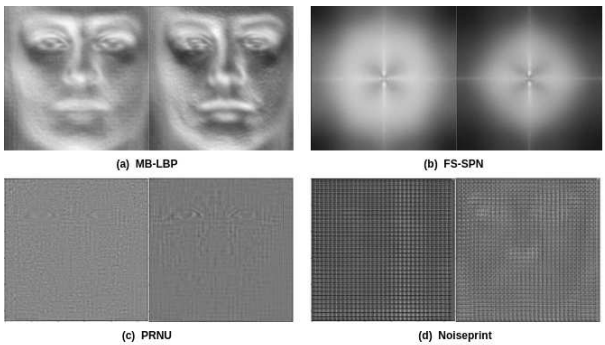}
\caption{Average of (a) MB-LBP, (b) FS-SPN, (c) PRNU and (b) Noiseprint features over 1000 randomly selected face images. Left: bona fide; right: morphed faces.}
\label{fig:featurefig}
\end{figure}

\par We first compare different feature-level fusion strategies to combine PRNU and Noiseprint patterns, including element-wise operation (sum/max), convex compression (CC) \cite{norouzi2013zero}, vector concatenation, and our factorized bilinear coding (FBC) method \cite{gao2020revisiting}. We consider the features in both the spatial and the spectral domains. The PRNU and Noiseprint features extracted from the images are treated as spatial features. The spectral features are obtained by applying the discrete Fourier transform to the spatial features. Any two types of feature are fused to perform traditional MAD tasks on a subset of the test data. Therefore, six different fusion features are generated. For concatenation, the final dimension of the feature is 145,800. For sum, max, and CC, it is 72,900. The fusion feature of FBC is as compact as 2,048 dimensions. All generated features are fed into the SVM with a linear kernel for binary classification. As shown in Table \ref{tab:toyexp}, the fusion of spatial features of PRNU and Noiseprint performs best for the six features, which can be attributed to the fact that the two patterns in the spatial domain contain more discriminative features (as shown in Fig.~\ref{fig:featurefig}). Furthermore, our FBC-based fusion achieves the highest accuracy among the five fusion strategies.

\begin{table*}[!t]
\small
\caption{Accuracy(\%) of 1-shot MAF classification on single and hybrid datasets.}
\vspace{-0.3cm}
\label{tab:maf1}
\resizebox{\linewidth}{!}{
\begin{tabular}{ l |c c c c c c}
\hline 
\multirow{2}{*}{Method} & FERET-Morphs & FRGC-Morphs & FRLL-Morphs & CelebA-Morphs & Doppelgänger & Hybrid \\
& 4-class & 4-class & 7-class & 4-class & 4-class & 9-class \\
\hline 
Xception \cite{chollet2017xception} & 29.47&	25.26&	17.68&	16.67&	21.05&	15.11\\
MobileNetV2 \cite{sandler2018mobilenetv2} & 31.58&	33.68&	31.30&	55.19&	25.26&	17.33\\
NasNetMobile \cite{zoph2018learning} & 32.63&	27.37&	22.61&	19.26&	23.16&	12.88\\
DenseNet121 \cite{huang2017densely} & 46.32&	26.32&	22.03&	47.04&	23.16&	19.33\\
ArcFace \cite{deng2019arcface} & 29.33&	39.64&	26.12&	28.33&	18.03&	15.22\\
\hline
Raghavendra. et al. \cite{raghavendra2017face} & 38.95&	43.16&	29.28&	89.63&	31.58&	11.11 \\ 
MB-LBP \cite{scherhag2020face} & 33.95 &	33.42&	34.59&	34.50 &	21.31&	14.89 \\
FS-SPN \cite{zhang2018face} & 25.41&	31.22&	23.71&	61.50 &	32.79&	29.44 \\	
Pipeline Footprint \cite{neubert2018reducing} & 26.32&	29.47&	29.28&	25.93&	25.26&	21.89 \\
PRNU Analysis \cite{debiasi2018prnu} & 34.74 & 26.32 & 11.01 & 37.04 &	25.26 &	18.56 \\
Inception-MAD \cite{damer2022privacy} & 23.16 &	30.53 &	20.00	& 44.81 & 29.47 & 21.78 \\
MixFaceNet-MAD \cite{damer2022privacy} & 36.84 & 37.89 & 35.94 & 57.04 & 49.47 & 33.56 \\
Noiseprint-SVM \cite{cozzolino2018noiseprint} & 50.53 & 43.16 & 22.61 & 84.44 & 31.58 & 22.00 \\
\hline
Meta-Baseline \cite{chen2021meta} & 51.05 & 51.44 & 34.77 & 61.43 & 33.43 & 53.46 \\
COSOC \cite{luo2021rectifying} & 54.58 & 64.37 & 35.22 & 63.19 & 34.30 & 59.55 \\
\hline
FBC & 96.93 & 98.83 & 94.06 & 99.50 & 56.67 & 96.11 \\
FBC-all & 98.11 & 99.48 & 98.42 & 100 & 84.17 & 96.78 \\
\textbf{FBC-APL} & \textbf{98.82} & \textbf{99.61} & \textbf{98.24} & \textbf{99.67} & \textbf{91.67} & \textbf{98.11} \\
\hline
\end{tabular}}
\end{table*}

\begin{table*}[!t]
\small
\caption{Accuracy(\%) of 5-shot MAF classification on single and hybrid datasets.}
\vspace{-0.3cm}
\label{tab:maf5}
\resizebox{\linewidth}{!}{
\begin{tabular}{ l |c c c c c c}
\hline 
\multirow{2}{*}{Method} & FERET-Morphs & FRGC-Morphs & FRLL-Morphs & CelebA-Morphs & Doppelgänger & Hybrid \\
& 4-class & 4-class & 7-class & 4-class & 4-class & 9-class \\
\hline 
Xception \cite{chollet2017xception} & 46.32& 43.16&	31.01&	73.70&	28.42&	43.67\\
MobileNetV2 \cite{sandler2018mobilenetv2} & 55.79 & 53.68 &	40.00 & 89.26 & 26.32 & 54.56 \\
NasNetMobile \cite{zoph2018learning} & 48.42 & 40.00 & 24.35 &	67.41&	27.37&	37.33\\
DenseNet121 \cite{huang2017densely} & 54.74 & 55.79 &	36.23&	89.26&	25.26	&53.33\\
ArcFace \cite{deng2019arcface} & 44.34 & 50.91 & 33.81 & 39.67 & 20.49 & 29.11 \\
\hline
Raghavendra. et al. \cite{raghavendra2017face} & 45.26 & 61.05 & 31.59 & 42.96 &	28.42 &	11.11 \\
MB-LBP \cite{scherhag2020face} & 69.28 & 74.87 & 42.67 & 63.00	& 26.23 & 42.11 \\
FS-SPN \cite{zhang2018face} & 41.34 & 41.97 & 26.91 & 82.67 & 27.04 & 43.89 \\	
Pipeline Footprint \cite{neubert2018reducing} & 45.26 & 61.05 & 31.59 &	42.96 & 28.42 & 37.78 \\
PRNU Analysis \cite{debiasi2018prnu} & 53.68 & 32.63 & 29.86 & 78.15 & 26.32 & 39.22 \\
Inception-MAD \cite{damer2022privacy} & 50.53 &	51.58 &	37.39 &	82.59 &	29.47 & 44.00 \\
MixFaceNet-MAD \cite{damer2022privacy} & 63.16	& 63.68 & 53.48 & 82.59 & 33.68 & 51.00 \\
Noiseprint-SVM \cite{cozzolino2018noiseprint} & 69.47 & 69.47 & 57.39  & 87.41 & 37.89 & 51.89 \\
\hline
Meta-Baseline \cite{chen2021meta} & 60.60 & 64.72 & 50.74 & 81.42 & 36.80 & 61.98 \\
COSOC \cite{luo2021rectifying} & 65.98 & 75.04 & 54.90 & 89.60 & 41.81 & 72.62 \\
\hline
FBC & 97.64 & 99.09 & 96.94 & 99.50 & 65.83 & 96.22 \\
FBC-all & 98.11 & 99.48 & 98.42 & 100 & 84.17 & 96.78 \\
\textbf{FBC-APL} & \textbf{98.82} & \textbf{99.61} & \textbf{98.24} & \textbf{99.67} & \textbf{96.67} & \textbf{98.22} \\
\hline
\end{tabular}
}
\end{table*}

\begin{figure*}[h]
\centering
\includegraphics[width=1.0\linewidth]{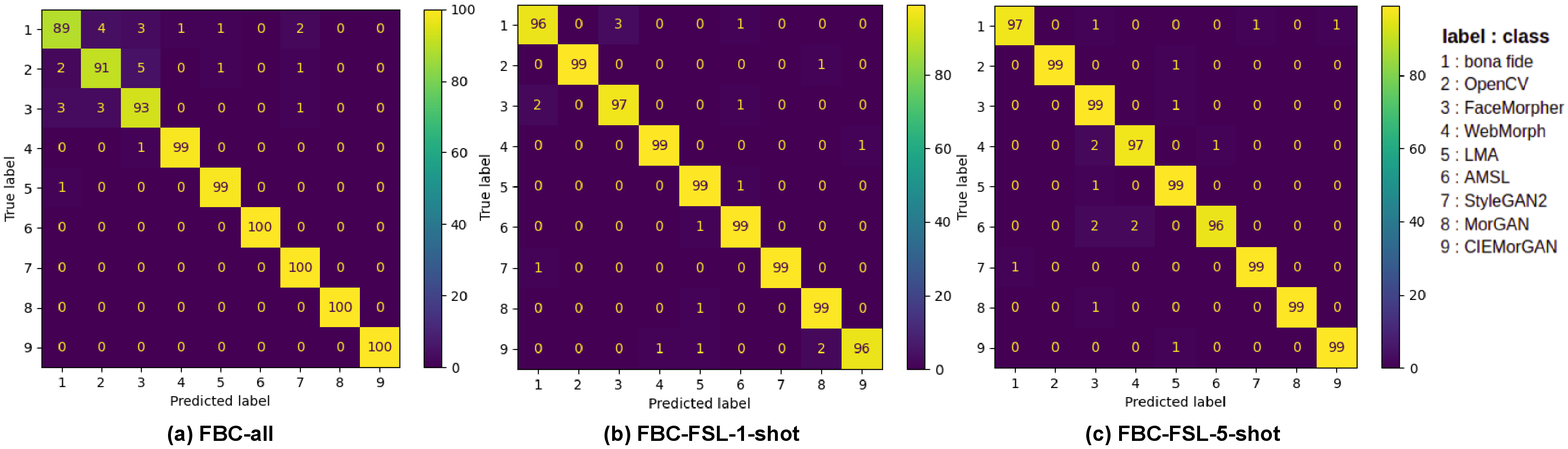}
\vspace{-0.75cm}
\caption{Confusion matrix of few-shot MAF classification on hybrid dataset.}
\label{fig:confus}
\end{figure*}

\subsection{Few-shot Learning for MAD}


We extend the traditional MAD problem to a few-shot learning problem. First, the PRNU and Noiseprint features are extracted, respectively. Then an FBC module (VGG-16 \cite{simonyan2014very} as the backbone) is trained as a binary classifier for feature fusion, taking PRNU and Noiseprint features from the entire training set (all images of predefined types) as input. Based on the pre-trained FBC module, 2,048-dimensional fusion representations are generated and then fed to the APL module for binary few-shot learning using the cross-entropy loss. Here, the Euclidean distance is used to query the top five nearest neighbors of the memory component. The APL output is a tuple of the probability distribution for each class. The results in terms of accuracy, D-EER, and ACER are shown in Table \ref{tab:madfs}. Two methods based on FSL \cite{luo2021rectifying,chen2021meta}, two methods based on face recognition (FR) \cite{schroff2015facenet,deng2019arcface}, several popular deep models pre-trained \cite{chollet2017xception,sandler2018mobilenetv2,zoph2018learning,huang2017densely} on ImageNet \cite{deng2009imagenet}, and eight current MAD methods \cite{raghavendra2017face,scherhag2020face,zhang2018face,neubert2018reducing,debiasi2018prnu,damer2022privacy,cozzolino2018noiseprint}, are adopted for comparison. Due to the effective fusion of two complementary patterns (i.e., PRNU and Noiseprint) and the APL module, our proposed FBC-APL clearly outperforms other competing methods by a large margin. 

\subsection{Few-shot Learning for MAF}
Unlike the few-shot MAD problem, in MAF, the FBC module uses ResNet50 \cite{he2016deep} as the backbone and is pre-trained as a nine-class classifier using all the training data (about 80\%) of the collected database. The FBC fusion feature obtained from the training samples is then fed to the APL module for multiclass few-shot learning. A cosine similarity score is adopted to compute the similarity between queries and the data stored in memory to find the three nearest neighbors. From Tables \ref{tab:maf1} and \ref{tab:maf5}, one can see that our FBC-APL has achieved outstanding performance, and some results are even better than the FBC-all method, which uses FBC features from all training data to fit SVM for classification. To better illustrate the effectiveness of the proposed FBC-FSL method, we have compared the confusion matrix for nine different classes (including bona fide and eight different morphing models), as shown in Fig. \ref{fig:confus}. 

\subsection{Discussions and Limitations}
Why did the proposed method outperform other competing methods by a large margin? We believe there are three contributing reasons. First, PRNU and Noiseprint feature maps as shown in Fig. \ref{fig:featurefig} have shown better discriminative capability than others; meanwhile, their complementary property makes fusion an efficient strategy for improving the accuracy. Second, we have specifically taken the few-shot constraints into the design (i.e., the adoption of APL module) while other competing approaches often assume numerous training samples. Third, from binary MAD to multi-class MAF, our FBC fusion strategy is more effective on distinguishing different classes as shown in Fig. \ref{fig:confus}. Note that we have achieved unanimously better results than other methods across six different datasets, as shown in Table \ref{tab:maf5}, which justifies the good generalization property of our approach.

The overall pipeline in Fig. \ref{fig:pipeline} can be further optimized by end-to-end training. In our current implementation, the three steps are separated, that is, the extraction of PRNU and Noiseprint features, FBC-based fusion, and APL-based FSL. From the perspective of network design, end-to-end training could further improve the performance of the FBC-APL model. Moreover, there are still smaller and more challenging datasets for morphing attacks in the public domain. Validation of the generalization property for the FBC-APL model remains to be completed, especially when novel face morphing attacks (e.g., adversarial morphing attack \cite{wang2020amora}, transformer-based, and 3D reconstruction-based face morphing) are invented. Finally, we have not considered the so-called post-morphing process \cite{damer2021pw} where the print and scan operations are performed when issuing a passport or identity document.

\section{Conclusion and Future Work}
\label{con}
\par Face morphing attacks pose a serious security threat to FRS. In this work, we proposed a few-shot learning framework for the detection of non-reference morphing attacks and fingerprinting problems based on factorized bilinear coding of two types of camera fingerprint feature, PRNU and Noiseprint. Additionally, a large-scale database is collected that contains five types of face dataset and eight different morphing methods to evaluate the proposed few-shot MAD and fingerprinting problem. The results show outstanding performance of the proposed fusion-based few-shot MAF framework on our newly collected large-scale morphing dataset. 
We note that face-morphing attack and defense research is likely to coevolve in the future. Future work on the attack side will include the invention of more powerful morphing attacks, such as GANformer-based \cite{hudson2021generative} and diffusion model-based \cite{dhariwal2021diffusion}. Consequently, defense models that include MAD and MAF could focus on the study of the feasibility of detecting novel attacks and morphed face images from printed and scanned image data. In practical applications, optimizing differential morphing attack detection with live trusted capture is also an interesting new research direction.

\section*{Acknowledgments}
This work was partially supported by the NSF Center for Identification (CITeR) awards 20s14l and 21s3li.



\bibliographystyle{IEEEtran}
\bibliography{Fusion-based_Few-Shot_Morphing_Attack_Detection_and_Fingerprinting}


\end{document}